\documentclass{article}

\usepackage{microtype}
\usepackage{graphicx}
\usepackage{subcaption}
\usepackage{booktabs}
\usepackage{enumitem}

\usepackage{hyperref}

\usepackage[preprint]{paperstyle}

\usepackage{amsmath}
\usepackage{amssymb}
\usepackage{mathtools}
\usepackage{amsthm}

\usepackage[capitalize,noabbrev]{cleveref}

\theoremstyle{plain}

\theoremstyle{definition}

\theoremstyle{remark}

\usepackage[textsize=tiny]{todonotes}
\icmltitlerunning{Inter-Request Caching for Video DiT Serving}

\begin{document}

\twocolumn[
\begin{center}
{\LARGE\bfseries Beyond Few-Step Inference: Accelerating Video Diffusion Transformer Model Serving with Inter-Request Caching Reuse\par}
\vspace{0.8em}

{\large
Hao Liu$^{1}$ \quad
Ye Huang$^{1}$ \quad
Chenghuan Huang$^{2}$ \quad
Zhenyi Zheng$^{1}$ \\
Jiangsu Du$^{1,*}$ \quad
Ziyang Ma$^{2}$ \quad
Jing Lyu$^{2}$ \quad
Yutong Lu$^{1}$\par}

\vspace{0.5em}
{\normalsize $^{1}$Sun Yat-sen University, China \quad $^{2}$Independent Researcher\par}

{\normalsize \texttt{\{liuh393,huangy783,zhengzhy37\}@mail2.sysu.edu.cn} \quad \texttt{\{caderhuang,ziyangma,eckolv\}@tencent.com} \quad \texttt{\{dujiangsu,luyutong\}@mail.sysu.edu.cn}\par}
\end{center}
\vspace{0.2em}
]
\begingroup
\renewcommand\thefootnote{}
\footnotetext{\hspace*{-1.5em}$^{*}$ Corresponding author: dujiangsu@mail.sysu.edu.cn.}
\endgroup

\makeatletter\global\icml@noticeprintedtrue\makeatother

\begin{abstract}
Video Diffusion Transformer (DiT) models are a dominant approach for high-quality video generation but suffer from high inference cost due to iterative denoising.
Existing caching approaches primarily exploit similarity within the diffusion process of a single request to skip redundant denoising steps.
In this paper, we introduce Chorus, a caching approach that leverages similarity across requests to accelerate video diffusion model serving.
Chorus achieves up to 45\% speedup on industrial 4-step distilled models, where prior intra-request caching approaches are ineffective.
Particularly, Chorus employs a three-stage caching strategy along the denoising process.
Stage 1 performs full reuse of latent features from similar requests.
Stage 2 exploits inter-request caching in specific latent regions during intermediate denoising steps.
This stage is combined with Token-Guided Attention Amplification to improve semantic alignment between the generated video and the conditional prompts, thereby extending the applicability of full reuse to later denoising steps.
Stage-3 operates in the final steps and disables caching reuse to repair discontinuities.
\end{abstract}

\section{Introduction}

Video Diffusion Transformer (DiT) models, such as Wan~\cite{wan2025wan} and Hunyuan~\cite{kong2024hunyuanvideo}, have substantially advanced the quality of video synthesis.
However, these improvements come at a significant computational cost.
For instance, the vanilla Wan2.1 takes 16.9 minutes to generate a 5-second video on a single NVIDIA A100 GPU.
As a result, video DiT models are prohibitively expensive for large-scale deployment and practical video generation services.

To improve computational efficiency, various acceleration techniques have been proposed~\cite{mhla, lin2024awq, zhang2025slasparsitydiffusiontransformers}, among which feature caching-based approaches~\cite{xu2018deepcache} have shown promise.
Video DiTs employ an iterative denoising process with tens of steps, during which highly similar features are observed across adjacent timesteps.
To this end, intra-request caching approaches~\cite{liu2025timestep, liu2025reusingforecastingacceleratingdiffusion} store features from previous timesteps and reuse them in subsequent ones, allowing diffusion models to skip redundant computations and reduce inference latency.
However, in addition to limited acceleration, it often fails for industrial 4-step distilled models, where intra-request redundancy has largely been removed.

\begin{figure}
    \centering
    \includegraphics[width=1\linewidth]{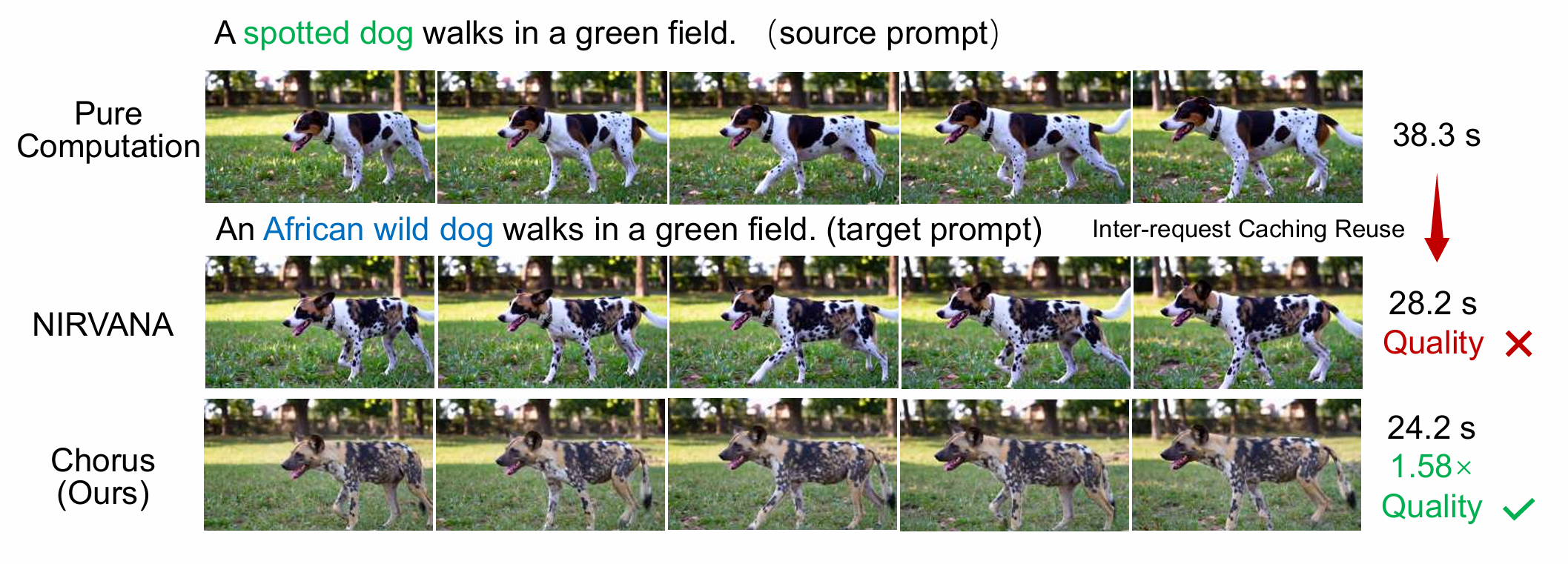}
    \caption{The video generation illustration of Chorus on a distilled 4-step Wan2.1 model. Chorus reuses cached latent states from a semantically similar request. It significantly accelerates inference while maintaining accurate semantic generation.}
    \vskip -0.1in
    \vspace{-8pt}
    \label{fig:introduction-pic}
\end{figure}

Furthermore, feature similarity across different requests has also been observed in DiT models, and such similarity tends to increase as requests accumulate.
This observation motivates inter-request caching approaches, particularly for applications with a well-defined scope.
For example, Adobe's NIRVANA~\cite{agarwal2024approximate} reports high similarity among prompt queries in its text-to-image service, where users often issue similar prompts within a session and prompts across different users also exhibit substantial overlap. 
As shown in Fig.~\ref{fig:introduction-pic}, this observation motivates exploiting inter-request similarity to accelerate its image DiT inference.
Although inter-request caching has proven effective in image generation, it remains a significant challenge in video generation. 

To leverage inter-request caching for video DiT serving, we propose Chorus.
It adopts a three-stage inter-request caching strategy throughout the denoising process, consisting of \textbf{full reuse} in the initial stage (stage-1), \textbf{selective region reuse} in the intermediate stage (stage-2), and \textbf{full computation} in the final stage (stage-3).
Stage 1 follows a similar idea to NIRVANA by reusing latent features at a specific initial denoising step.
However, since video sequences are considerably longer than images and are subject to an additional temporal consistency constraint, naive caching reuse must start from very early denoising steps; otherwise, semantic inaccuracies may arise.
We propose \textbf{Token-Guided Attention Amplification (TGAA)}, a mechanism that strengthens the influence of the whole prompt and key tokens in subsequent denoising steps, thereby improving semantic alignment and enabling more steps to be safely skipped.

As denoising progresses, the generation becomes increasingly specific, making full reuse ineffective.
Accordingly, stage 2 performs selective region reuse, reusing latent states corresponding to specific objects or background regions.
Given that attention dominates the computation in video DiT models and its cost scales quadratically with sequence length, selective region reuse can further reduce overall computation by restricting attention to shorter sequences.

Finally, in the late steps, the generation becomes highly specific and sensitive to fine-grained details. At this stage, even region-based reuse may introduce visual or semantic discontinuities. To recover from such artifacts, Stage 3 disables reuse entirely and performs full computation, allowing the model to refine details and restore generation consistency.

Our contributions can be summarized as follows:
\vspace{-3mm}

\begin{itemize}[leftmargin=*]
    \item We propose Chorus, a three-stage inter-request caching approach that enables effective reuse throughout the video DiT denoising process. To the best of our knowledge, Chorus is the first work to effectively exploit inter-request caching in video DiT.
    \item We conduct extensive evaluations on both vanilla and industrial distilled DiT models, demonstrating that Chorus achieves significant acceleration while preserving generation quality. Moreover, Chorus is orthogonal to distillation and intra-request caching approaches, and can be combined to further improve performance beyond the state of the art.
\end{itemize}
\vspace{-2mm}

\begin{figure}
    \centering
    \includegraphics[width=0.95\linewidth]{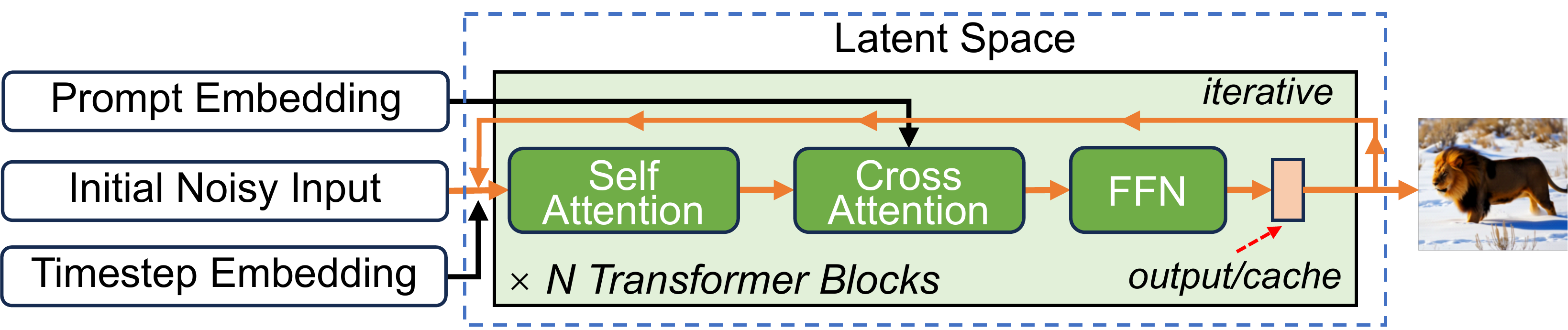}
    \caption{The DiT model architecture and inference process.}
    \label{fig:dit}
    \vspace{-8pt}
    \vskip -0.1in
\end{figure}

\section{Background}
\subsection{Diffusion Transformer Model}
Diffusion Transformer (DiT) models~\cite{peebles2023scalable-dit} integrate diffusion modeling~\cite{ho2020denoising,sohl2015deep} with Transformers~\cite{vaswani2017attention} and have emerged as a dominant paradigm for image~\cite{rombach2022high,dhariwal2021diffusion} and video generation~\cite{ho2022video}.
As shown in Fig.~\ref{fig:dit}, a DiT model generally consists of a Transformer backbone as the core component.
Moreover, since video is represented as an extremely long sequence, video DiT inference is typically performed with a batch size of one, and the attention module dominates the overall computational cost.

Generation begins from a noise-initialized input, which is progressively refined by the Transformer through a sequence of denoising steps.
At each step, prompt conditioning and timestep embeddings are injected into the model input to steer the denoising process.
Notably, the prompt conditioning is used to guide the denoising process by cross attention.
The output produced at one step is then fed back as the input to the next, forming an iterative refinement loop until the final result is obtained.

\subsection{Distilled DiT Models}
Knowledge distillation~\cite{hinton2015distilling} is a model compression technique that transforms complex models into lighter and more efficient ones while largely preserving performance.
For video DiT models, distillation can reduce the number of sampling timesteps while maintaining output quality~\cite{salimans2022progressive-distill,song2023consistency-distill,yin2024improved-distillDMD2}, thereby significantly alleviating the computational bottleneck.
In industrial scenarios, distilled video DiT models~\cite{imageteam2025zimageefficientimagegeneration,zhang2025turbodiffusionacceleratingvideodiffusion} are commonly adopted with 4–8 sampling steps, achieving up to \textbf{26$\times$} acceleration compared to standard 50-step models.

\subsection{Intra-request Caching Reuse}
During the denoising process, intermediate outputs at adjacent timesteps often exhibit high similarity, particularly in later steps.
As illustrated in Fig.~\ref{fig:intra-inter-reuse}, intra-request caching reuse exploits this temporal redundancy within a single request by skipping some steps.
Early methods, such as TeaCache~\cite{liu2025timestep} and PAB~\cite{kahatapitiya2025adaptive-intra-request}, focus on directly reusing cached outputs from previous timesteps, whereas more recent approaches, including TaylorSeer~\cite{liu2025reusingforecastingacceleratingdiffusion} and Hiten~\cite{feng2025hicache-intra-request}, further extend this idea by predicting features for the current timestep using learned functions.
These intra-request reuse techniques can achieve significant speedups, particularly for vanilla video DiT models.
Unfortunately, these approaches are not orthogonal to distillation techniques, as both primarily eliminate redundancy across denoising steps.
As a result, intra-request caching reuse methods often become ineffective on practical distilled models.
They cannot further reduce the number of denoising steps.

\subsection{Inter-request Caching Reuse}

\begin{figure}
    \centering
    \includegraphics[width=0.95\linewidth]{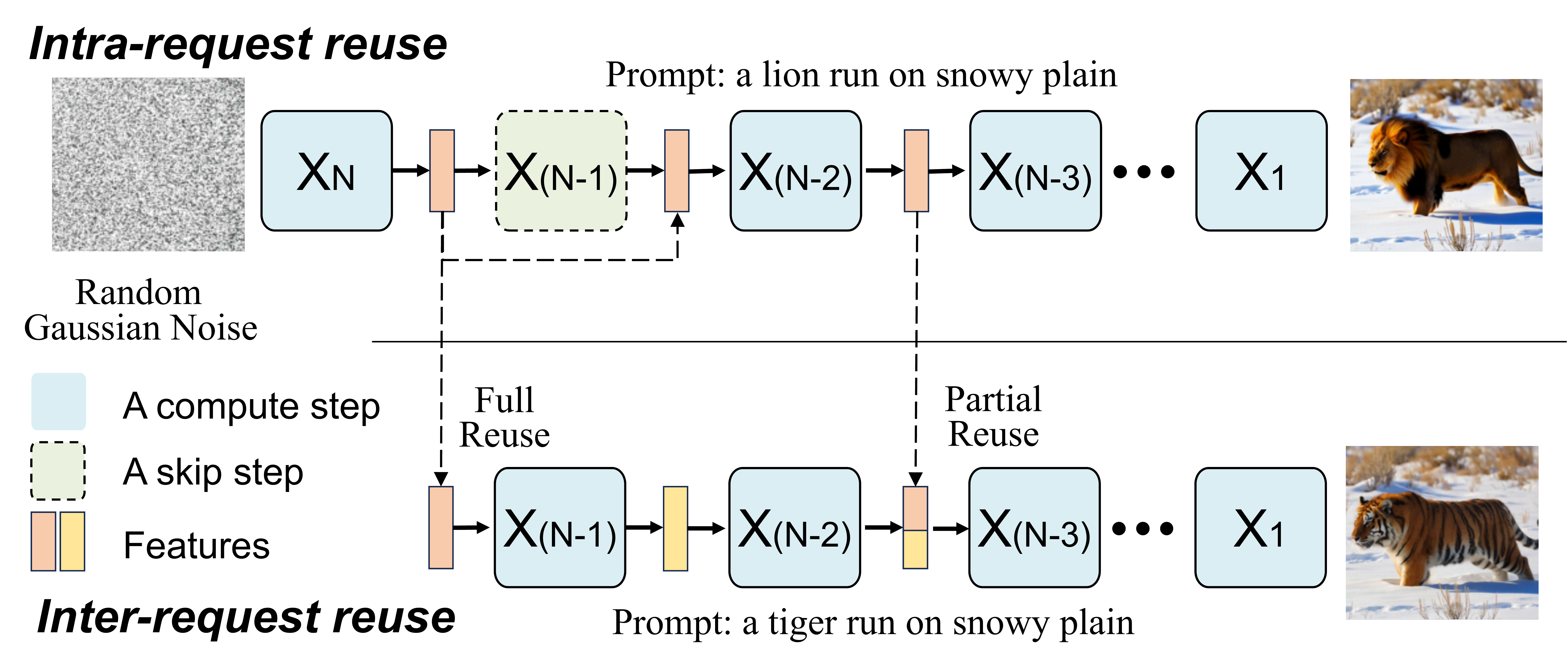}
    \vspace{-2mm}
    \caption{Illustration of intra- and inter-request caching reuse, with denoising timesteps explicitly expanded.}
    \label{fig:intra-inter-reuse}
    \vspace{-2mm}
\end{figure}

Besides similarity between timesteps of a single request, the inter-request caching reuse leverages the similarity across requests.
For instance, Adobe's NIRVANA~\cite{agarwal2024approximate} observes that feature similarity across requests increases as the number of requests accumulates and exploits this property for image generation.
As shown in Fig.~\ref{fig:intra-inter-reuse}, when a new request arrives, it does not generate images from scratch.
Instead, it identifies a matched intermediate output at a corresponding denoising step from another request and continues generation from that state, thereby skipping several initial denoising steps.
We refer to this inter-request mechanism as full reuse.

Since videos are significantly larger than images and involve much longer sequences, inter-request reuse becomes more sensitive and can only be safely applied from very early denoising steps.
In Chorus, we introduce a novel mechanism to enable full reuse that skips more denoising steps, and further propose partial reuse, as in Fig.~\ref{fig:intra-inter-reuse}, to selectively skip computation in intermediate steps.

\vspace{-1mm}
\section{Chorus Design}
\subsection{Overview of the Chorus Framework}

\begin{figure}
  \begin{center}
    \centerline{\includegraphics[width=0.98\linewidth]{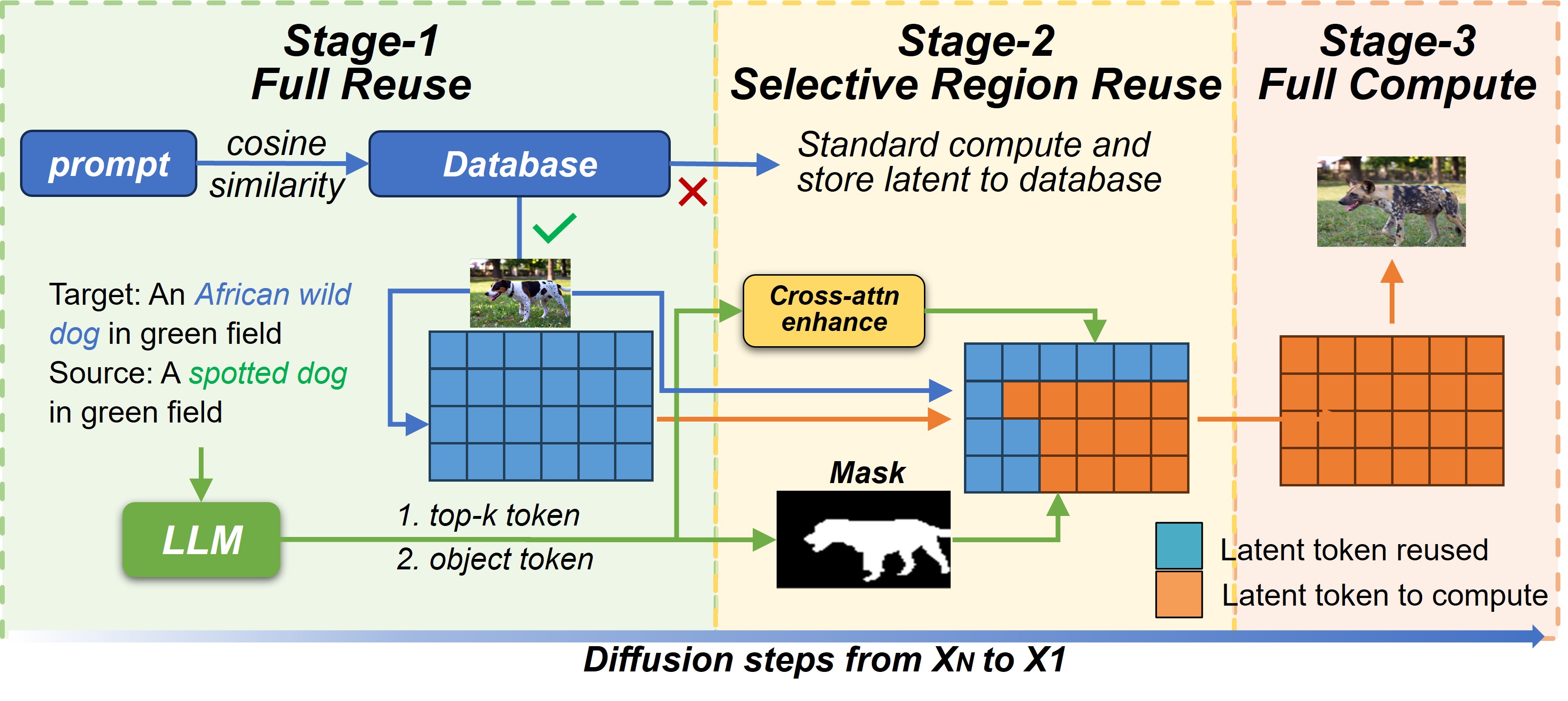}}
    \vspace{-1mm}
    \caption{Chorus Overview.}
    \vspace{-8mm}
    \label{fig:overview}
  \end{center}
\end{figure}

The video generation process in Chorus is illustrated in Fig.~\ref{fig:overview}.
Chorus first overcomes the challenge that existing full inter-request reuse approaches can only be applied at very early denoising steps.

By introducing a Token-Guided Attention Amplification (TGAA) mechanism, Chorus preserves semantic alignment while enabling full reuse across substantially more denoising steps.
Next, to overcome the challenge that only the initial stage can benefit from the inter-request caching, Chorus proposes to reuse caching from other requests in a region-based manner.
In this way, the video generation process in Chorus is naturally divided into 3 stages. For a target prompt:

\vspace{-4mm}

\begin{itemize}[leftmargin=*]
    \item Full Reuse (Stage-1). The system retrieves cached source requests whose CLIP embeddings~\cite{radford2021learning} exhibit the highest cosine similarity to that of the target prompt. If the similarity exceeds a threshold $\tau$, the target and source prompts are considered semantically similar. Upon a successful match, the generation process reuses the latent features from the matched source prompt for a predefined number of denoising steps. Otherwise, the system falls back to standard generation.
    \item Selective Region Reuse (Stage-2).
    To extend inter-request reuse to subsequent steps, we selectively compute regions that differ between the target and source prompts while reusing latent features for semantically similar regions.

    To identify prompt-differentiated regions, we employ a lightweight LLM to extract object tokens from the text prompt and then map these tokens to visual regions using a segmentation model~\cite{ren2024groundedsamassemblingopenworld}.
    In addition, TGAA is applied in Stage 2 to further improve semantic alignment, where the lightweight LLM is used to identify differentiating tokens that guide attention enhancement.
    \item Full Compute (Stage-3). In the final denoising steps, we perform full computation to repair any discontinuities introduced by mask-driven acceleration across different regions and to refine fine-grained details.
\end{itemize}

\vspace{-1mm}
\textbf{Stage Boundary Determination.} Suppose the denoising process consists of $N$ steps. Chorus partitions the denoising process into three stages by determining two transition points, $K_1$ and $K_2$, where $0 \leq K_1 \leq K_2 \leq N$. 
These stage boundaries are determined by a scheduler via a monotonic mapping function $\mathcal{F}$ parameterized by a matching score $m$:
\begin{equation}
\label{match2}
(K_1, K_2) = \mathcal{F}(m; N).
\end{equation}
The monotonicity of $\mathcal{F}$ ensures that stronger inter-request similarity leads to later transition points, enabling more aggressive reuse while preserving semantic alignment.
In other words, higher similarity leads to larger values of $K_1$ and $K_2$. If no cached prompt meets the minimum similarity threshold $\tau$ (i.e., $m < \tau$), we set $K_1 = K_2 = 0$.

\textbf{Cache Update.} If no suitable cached request is found (i.e., $m < \tau$), the system falls back to standard generation. In this case, the intermediate latent states, prompt, and generated video of the request are stored in the cache for potential reuse by future requests.

\begin{figure}
    \centering
    \includegraphics[width=0.9\linewidth]{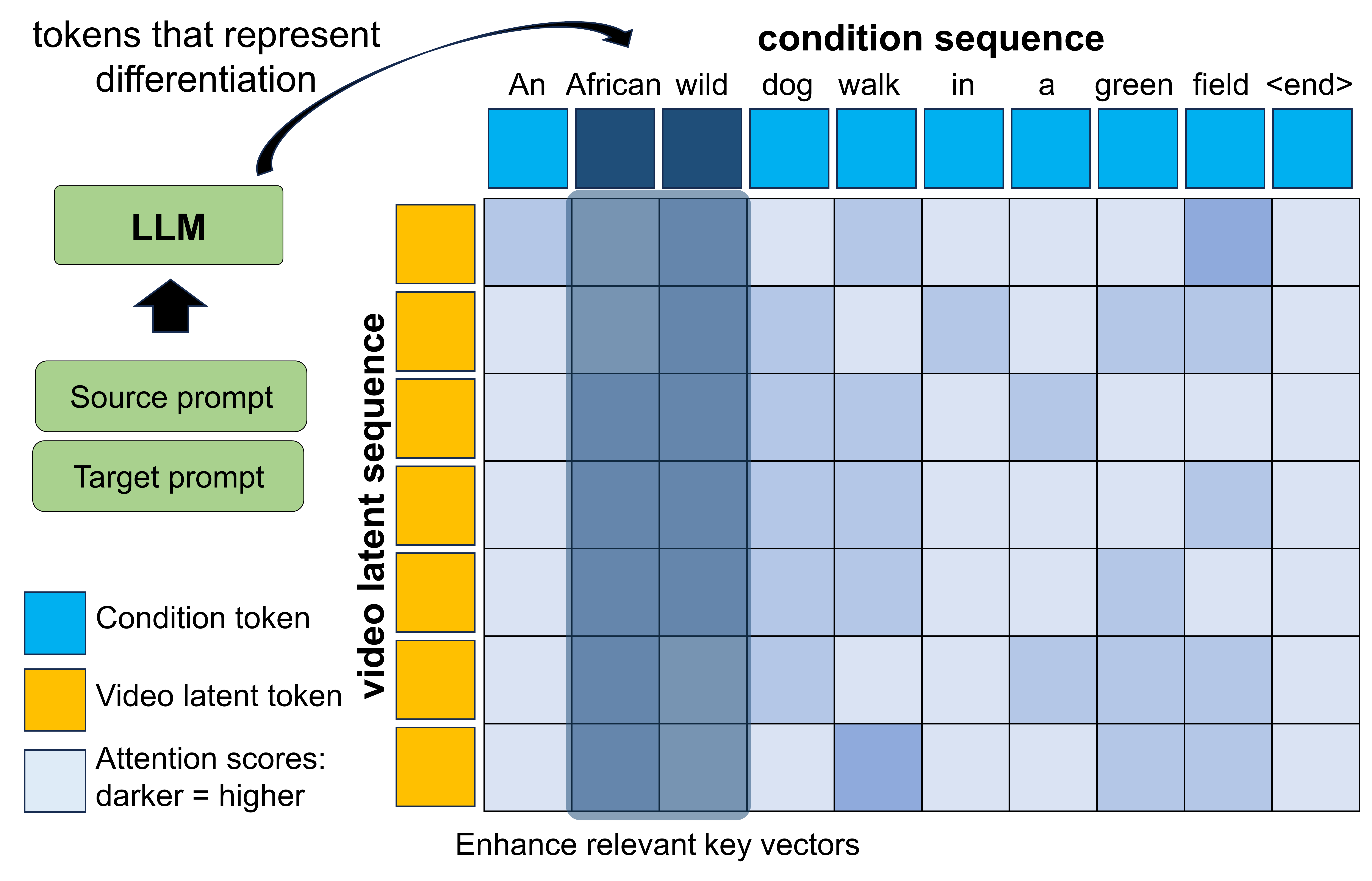}
    \vspace{-2mm}
    \caption{The illustration of key amplification in TGAA.We dynamically amplify key vectors of condition tokens (e.g., “African”, “wild”) that best capture the semantic difference in cross-attention operator to guide video latents.}
    \label{fig:tgaa}
    \vspace{-5mm}
\end{figure}

\subsection{Token-Guided Attention Amplification (TGAA)}

When extending previous inter-request latent reuse method from image to video generation, we observe persistent semantic misalignment: reused video latents tend to preserve attributes of the source prompt instead of fully adapting to the target prompt.
For example, reusing latents from “a spotted dog” to generate “an African wild dog” often yields videos that retain the spotted appearance, as shown in Fig.~\ref{fig:introduction-pic}.

We attribute this post-reuse misalignment primarily to the constraint of temporal consistency prior in the model. While essential for inter-frame coherence, the strong temporal prior inherently restricts the flexibility of latent updates. This prior functions as a strict regularizer that suppresses the significant feature shifts required for semantic alignment with the target prompt, thereby inhibiting the generation of visual features aligned with the target semantics.

To address the misalignment between reused video latents and target prompt conditions, we propose \textbf{Token-Guided Attention Amplification(TGAA)}. This mechanism adaptively identifies tokens exhibiting the most significant semantic shifts and selectively amplifies their influence to reinforce text guidance, thereby ensuring precise alignment with the target prompt.

TGAA operates in three steps: (1) identifying and boosting the key vectors of condition tokens that best capture the semantic difference;
(2) strengthening the entire cross-attention output; and (3) dynamically decaying the amplification strength across denoising process. 

\textbf{Key Amplification.}
TGAA first introduces precise semantic guidance by identifying and amplifying the key vectors of the condition tokens that capture the semantic differences between the target and source prompts. 
These differential tokens are identified by comparing the target and source prompts using a lightweight LLM.
As illustrated in Fig.~\ref{fig:tgaa}, during cross-attention, the key vectors corresponding to these tokens are selectively scaled by a factor \(\gamma_k\):
\begin{equation}
    \begin{aligned}
    & K'[:, \mathcal{T}_{\text{diff}}, :] = \gamma_k \cdot K[:, \mathcal{T}_{\text{diff}}, :],
\quad K \in \mathbb{R}^{B \times L' \times d_k} \\
    & \text{Attention}(Q, K', V) = \text{softmax}\left( \frac{Q K'^{\top}}{\sqrt{d_k}} \right) V
    \end{aligned}
\end{equation}

where \(\mathcal{T}_{\text{diff}}\) denotes the indices of the differential tokens.  
This targeted key amplification sharpens the attentional focus of the latent queries on the new semantic attributes. 
By strengthening the key projections of the differential tokens, the resulting attention scores are skewed toward those tokens, which effectively forces the latent queries to prioritize the newly introduced attributes, thereby driving the semantic transformation more effectively.

\textbf{Output Amplification.} TGAA simultaneously applies a global scaling factor to the entire cross-attention output to reinforce prompt conditioning. In a standard DiT block, the cross-attention output is added to the latent features and we add a scaling factor \(\gamma_o\) to strengthen the influence of the text prompt on every latent token:
\vspace{-1mm}
\begin{equation}
\label{eq:output_scaling}
x_{\text{out}} = x_{\text{in}} + \gamma_o \cdot \text{cross\_attn}(x_{\text{in}}).
\end{equation}
\vspace{-4mm}

In summary, key amplification identifies the target-specific attributes that define the desired update direction, while output amplification globally strengthens the conditioning signal to drive the latent sequence more strongly along that direction. They work together to guide the denoising updates toward the target prompt.

\begin{figure}

    \includegraphics[width=0.9\linewidth]{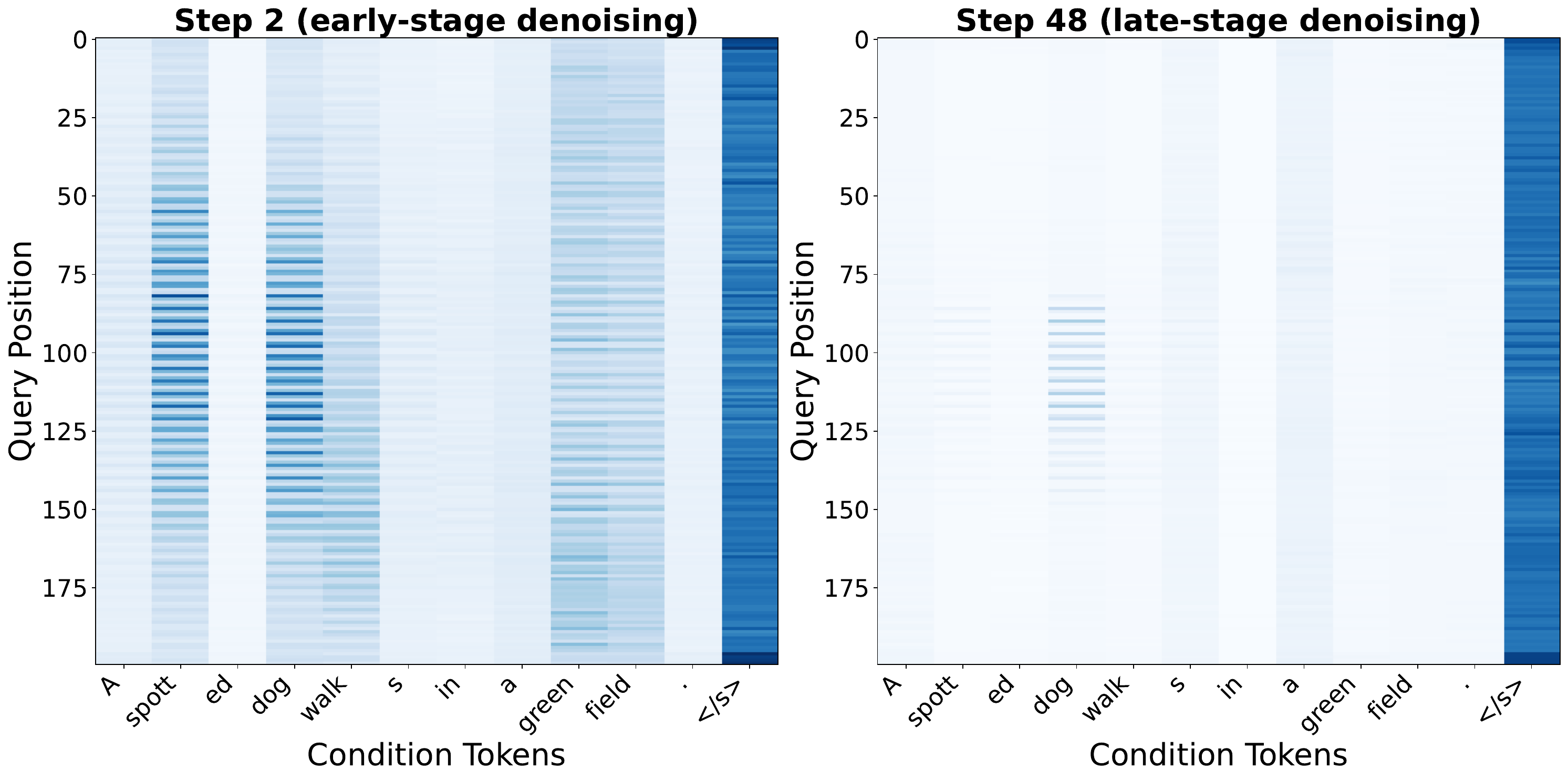}

    \vspace{-1mm}
    \caption{Attention map of cross-attention. Cross-attention is more influential in the early denoising stages, while its effect diminishes later, with nearly all attention scores concentrating on the EOS token. We visualize step 2 (early-stage denoising) and step 48 (late-stage denoising).}
    \vspace{-5mm}
    \label{fig:cross_attention_map}
\end{figure}

\textbf{Dynamic Decay Scheduler.} Although attention amplification is effective, it is not feasible to keep two amplification factors $\gamma_k$ and $\gamma_o$ constant throughout the denoising process.
As shown in Fig.~\ref{fig:cross_attention_map}, text-condition guidance is most effective during the early steps for establishing the overall structure and macro properties of the video, while its influence diminishes in later steps as its attention scores become dominated by the EOS token.
A static, non‑adaptive amplification strategy mismatches this dynamic, leading to distinct artifacts: excessive output scaling blurs frames, while over‑amplifying specific keys suppresses other critical tokens (e.g., motion words), harming temporal coherence. 

Therefore, we design a dynamic decay scheduler that adapts the amplification strength according to the denoising progress. 
The scheduler adjusts \( \gamma_k(t) \) and \( \gamma_o(t) \) as functions of the current timestep \( t \) and the matching score \( m \):
\vspace{-2mm}
\begin{equation}
\gamma_k(t) = F_k(t, m), \qquad \gamma_o(t) = F_o(t, m)
\end{equation}
\vspace{-5mm}

The modulation functions \( F_k, F_o \) are designed to satisfy three key principles:
\vspace{-2mm}
\begin{itemize}[leftmargin=*]
    \item The factors decrease as denoising progresses, providing stronger guidance in early steps while gradually weakening the influence in later steps.
    \item For a large matching score (\( m \)), the amplification is reduced to avoid over‑forcing unnatural transformations.
    \item The factors are clamped to be no smaller than $1.0$, ensuring that the conditioning signal is never weakened.
\end{itemize}

\subsection{Selective Region Denoising (SRD)}

\begin{figure}
    \centering
    \includegraphics[width=0.95\linewidth]{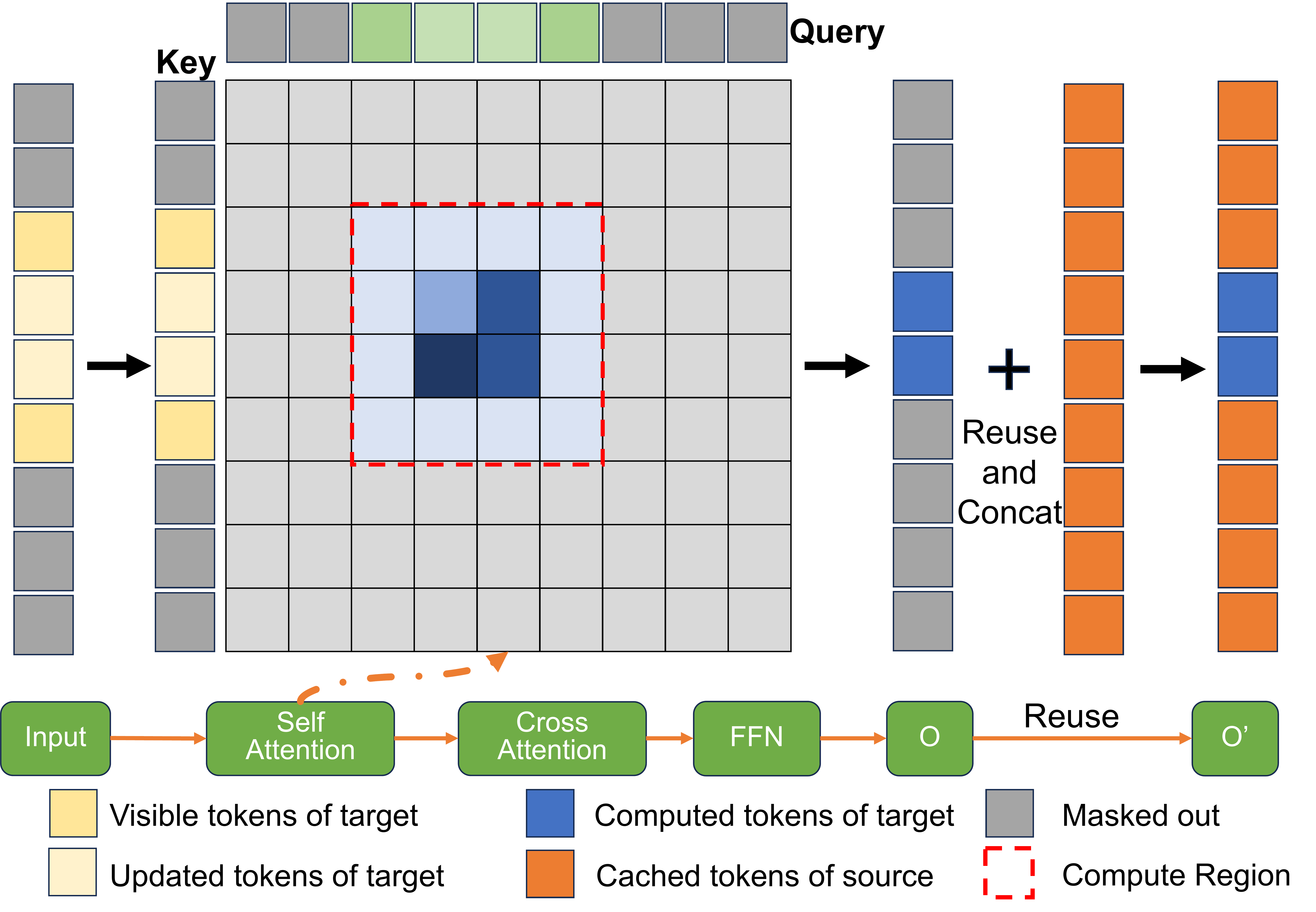}
    \vspace{-1mm}
    \caption{The illustration of selective region denoising.}
    \vspace{-5mm}
    \label{fig:SRD-show}
\end{figure}

Despite TGAA improving semantic alignment and enabling full reuse across more denoising steps, the remaining denoising steps still constitute the majority of the process, limiting the overall impact of inter-request caching reuse.
As the generation becomes increasingly specific, full latent reuse, even with TGAA, often leads to significant visible quality degradation and semantic drift in regions that differ between prompts.
We observe that relying solely on full reuse and extending it to later denoising steps is overly coarse.
In many cases, semantic mismatches are localized to specific regions, rather than affecting the entire generation.
For instance, transforming a “spotted dog” into an “African wild dog” allows background semantics to be reused, whereas updates are needed only in the object region.

Selective Region Denoising (SRD) adopts a fine-grained strategy to enable partial caching reuse and compute only a small set of regions.
Generally, it persistently reuses well-aligned regions, restricting recomputation exclusively to divergent regions.
As illustrated in Fig.~\ref{fig:SRD-show}, SRD first generates a region mask to identify divergent regions between the target and source prompts.
The mask is then applied to restrict computation in the self-attention, cross-attention, and FFN layers to only these divergent regions.
Finally, the newly computed regions are merged with the cached latent features from the source prompt to form the updated latent representation.

Thus, SRD necessitates addressing three core challenges: 
(i) accurately localizing divergent regions, 
(ii) ensuring spatial consistency between recomputed and reused areas, 
and (iii) effectively turning this computational sparsity into actual acceleration.

\begin{figure}
    \centering
    \includegraphics[width=1\linewidth]{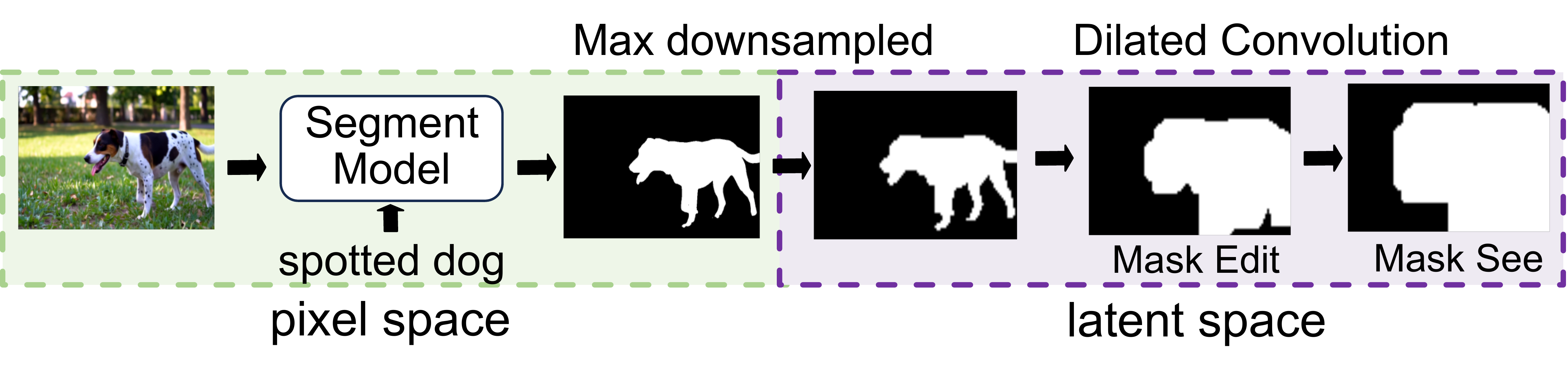}
    \vspace{-4mm}
    \caption{The mask generation process.}
    \label{fig:maskgen_show}
    \vspace{-5mm}
\end{figure}

\subsubsection{Region Mask Generation}
We use the lightweight LLM to compare target and source prompts and identify the divergent regions, and then generate a binary mask \( \mathbf{M}_{\text{base}} \) to distinguish where the source and target prompts differ. 
As shown in Fig.~\ref{fig:maskgen_show}, the mask generation pipeline consists of three stages:
\vspace{-1mm}
\begin{enumerate}[topsep=2pt,itemsep=2pt,parsep=2pt,leftmargin=*]
    \item \textbf{Object Prompt Extraction.} A LLM compares the source and target prompts to extract the most salient divergent object phrase (e.g., ``spotted dog'').
    \item \textbf{Pixel-Space Segmentation.} Using the extracted phrase as a text prompt, a pretrained segmentation model~\cite{ren2024groundedsamassemblingopenworld} processes the stored video of source prompt to obtain a pixel-level mask. To achieve efficiency, we adopt a key-frame strategy. Masks are generated only for sparsely-sampled key frames and are propagated to adjacent frames within the same temporal group.
    \item \textbf{Latent-Space Projection.} The pixel‑space binary mask is downsampled via a max pooling operator to the latent space, where denoising operates.
\end{enumerate}

\subsubsection{Hierarchical Masks for Latent Fusion}
Strictly denoising only within the masked region $\mathbf{M}_{\text{base}}$ would introduce hard boundaries, leading to visible seams, as it constrains object motion and lacks sufficient surrounding context.
To address this, we introduce a hierarchical mask that expands the receptive field while explicitly indicating the active update regions.
Thus, we generate two levels of expanded masks based on the fundamental mask.
The \(\mathbf{M}_{\text{base}}\) is expanded via morphological dilation to create two specialized masks:

\vspace{-3mm}
\begin{equation}
\begin{aligned}
\mathbf{M}_{\text{edit}} = \mathbb{I}(\mathbf{M}_{\text{base}} * \mathbf{K}_r > 0) \\
\mathbf{M}_{\text{see}} = \mathbb{I}(\mathbf{M}_{\text{base}} * \mathbf{K}_{r'} > 0)
\end{aligned}
\end{equation}
\vspace{-2mm}

with \(r' > r\), where \(*\) is convolution with an all-ones kernel \(\mathbf{K}\), and \(\mathbb{I}\) is the indicator function. This yields the containment hierarchy \(\mathbf{M}_{\text{see}} \supseteq \mathbf{M}_{\text{edit}} \supseteq \mathbf{M}_{\text{base}}\). 
Particularly, \textbf{Edit Mask (\(\mathbf{M}_{\text{edit}}\))} defines the active update region. 
Its moderate dilation (\(r\)) accommodates object motion and transformation, ensuring all edits remain within this buffer zone. \textbf{Visible Mask (\(\mathbf{M}_{\text{see}}\))} defines the context window for attention computation.
Its larger dilation (\(r'\)) provides sufficient surrounding background, enabling the model to integrate the updated region with the reused region.

With these two masks, updates are strictly confined to the $\mathbf{M}{\text{edit}}$ region, preserving the fidelity of reused content.
Crucially, although feature updates are limited to $\mathbf{M}{\text{edit}}$, the attention layers are allowed to attend over the larger $\mathbf{M}_{\text{see}}$ region, providing sufficient contextual information.
This dual-mask design maintains temporal and spatial consistency between the updated and reused regions. Formally, features are updated by:

\begin{equation}
\mathbf{L} \leftarrow \mathbf{M}_{\text{edit}} \odot \mathbf{L} + (1 - \mathbf{M}_{\text{edit}}) \odot \mathbf{SL}
\end{equation}
\vspace{-7mm}

\subsubsection{SRD Implementation}

Although computational sparsity is introduced, it remains difficult to translate this sparsity into actual acceleration.
Unlike prior approaches where masks merely zero out inactive regions while still participating in computation, Chorus leverages the hierarchical masks to explicitly select the regions to be computed in the FlashAttention kernel~\cite{dao2022flashattention}.
Notably, Chorus is compatible with FlashAttention technique.
After constructing the hierarchical masks, we feed both the latent embeddings and the masks into the Transformer block, where the masks are used to guide selective computation, as illustrated in Fig.~\ref{fig:SRD-show}.

Since attention dominates inference latency in video generation, contributing roughly 80\% of the runtime at 720p~\cite{fu2025slidingwindowattentiontraining}, and scales quadratically with sequence length, stage-2 effectively reduces its cost by restricting attention to selected regions.
Empirically, when 50\% of the region is visible, SRD achieves a 1.8$\times$ speedup in self-attention and a 1.5$\times$ end-to-end acceleration per stage-2 step, while preserving visual fidelity with no obvious quality degradation.

\section{Experiments}
\begin{table*}[htbp]
\centering

\caption{Quantitative comparison on Wan2.1 text-to-video models.
(0)/(1k) denote cold-start and warm-start caches, respectively.
\textbf{Hit Latency} represents the average latency of requests with cache hit.
The matching threshold (CLIP similarity) for both distilled models and 50-step models are 0.75 and 0.65 respectively.
We highlight the \textbf{best} and \underline{second best} entries.}
\resizebox{\linewidth}{!}{
\begin{tabular}{lccccccc}
\toprule
\textbf{Method} & \textbf{CLIP-SCORE↑} & \textbf{vbench-q↑} & \textbf{Latency(s)↓} & \textbf{Hit Latency(s)↓} & \textbf{Speedup↑} & \textbf{Speedup(hit)↑} & \textbf{Hit Rate}\\
\midrule
\multicolumn{8}{c}{\textbf{Wan2.1-14B-T2V-distilled (81 frames, 480P, 4steps, no cfg)}} \\
\midrule
\textbf{baseline}       & \textbf{31.9970}  & 0.9000 & 64.15 & - & 1.00x & 1.00x & -\\
\textbf{NIRVANA(0)}     & 31.3038            & 0.8993 & 56.75 & 46.37 & 1.13x & 1.38× & 42.6\%\\
\textbf{Chorus(0)}      & \underline{31.5006}& \underline{0.9005} & 55.91 & \underline{44.51} & 1.15x & \underline{1.44×} & 42.6\%\\
\textbf{NIRVANA(1k)}    & 31.0635           & 0.8991 & \underline{53.79} & 46.93 & \underline{1.19x} & 1.37× & 58.9\%\\
\textbf{Chorus(1k)}     & 31.3081           & \textbf{0.9006} & \textbf{52.09} & \textbf{44.15} & \textbf{1.23x} & \textbf{1.45x} &  58.9\%\\
\midrule

\multicolumn{8}{c}{\textbf{Wan2.1-14B-T2V (81 frames, 480P, 50steps)}} \\
\midrule
\textbf{baseline}                   & \underline{31.3764} & 0.8976 & 1683.1 & -                     & 1.00x & 1.00x & -\\
\textbf{NIRVANA(1k)}                & 29.6641 & 0.8962 & 1381.2 & 1277.88 & 1.22x                   & 1.32× & 58.0\%\\
\textbf{teacache(l=0.2)}            & \textbf{31.4445} & 0.8971 & \underline{1082.1} & -            & \underline{1.55x} & - & -\\
\textbf{Chorus(1k)}                 & 30.5598 & \textbf{0.8987} & 1339.2 & \underline{1206.69}      & 1.26x & \underline{1.39×} & 58.0\%\\
\textbf{Chorus(1k)+teacache(l=0.2)} & 30.4624 & \underline{0.8983} & \textbf{866.4} & \textbf{782.24} & \textbf{1.94x} & \textbf{2.15×} & 58.0\%
\\
\midrule

\multicolumn{8}{c}{\textbf{Wan2.1-1.3B-T2V (81 frames, 480P, 50steps)}} \\
\midrule
\textbf{baseline}                   & \textbf{30.3637} & \textbf{0.8954} & 319.01 & - & 1.00x & 1.00x & -\\
\textbf{NIRVANA(1k)}                & 29.0643 & \underline{0.8945} & 261.46 & 245.72 & 1.22x & 1.30× & 58.0\%\\
\textbf{teacache(l=0.2)}            & \underline{30.3580} & 0.8940 & \underline{114.68} & - & \underline{2.78x} & - & -\\
\textbf{Chorus(1k)}                 & 29.7352 & 0.8912  & 251.62 & \underline{226.08} & 1.27x & \underline{1.41×} & 58.0\%\\
\textbf{Chorus(1k)+teacache(l=0.2)} & 29.5147 & 0.8868 & \textbf{103.09} & \textbf{89.16} & \textbf{3.09x} & \textbf{3.58×} & 58.0\%\\
\bottomrule
\end{tabular}
}
\label{tab:main_experiments}
\vspace{-10pt}

\end{table*}

\subsection{Experimental Setup}
We summarize the key experimental settings below; additional details are provided in the appendix.~\ref{sec:appendix_experiments}.

\textbf{Model Configuration.} Our experiments were conducted on the widely-used Wan2.1-T2V model family~\cite{wan2025wan}, evaluating three distinct variants: the acclaimed 14B-parameter \textbf{Wan2.1-14B-T2V} for video generation, its lightweight counterpart \textbf{Wan2.1-1.3B-T2V} with reduced layers and embedding size, and the \textbf{four-step distilled variant from LightX2V}~\cite{lightx2v2025}. 

All models were evaluated using the DPM++ solver~\cite{lu2025dpm}, with all other sampling parameters kept at their default values. 

\textbf{Hardware.}
Unless otherwise noted, all main experiments are conducted on NVIDIA H20 96GB GPUs.
Ablation studies are run on NVIDIA A100 80GB GPUs.

\textbf{Dataset}. We evaluate Chorus on dataset VidProM ~\cite{wang2024vidprom-dataset}, which contains 1.67 M unique text-to-Video prompts from real-world users on Discord. 
To expedite evaluation, we sample 2{,}000 prompts from VidProM: 1{,}000 to initialize the cache and 1{,}000 for testing.

\textbf{Baselines and Cache Settings.}
We compare against the \textbf{no-reuse} baseline (vanilla inference), \textbf{TeaCache}~\cite{liu2025timestep} as an intra-request caching baseline, and \textbf{NIRVANA}~\cite{agarwal2024approximate} as an inter-request reuse baseline.
For inter-request methods (Chorus/NIRVANA), we evaluate two cache initialization settings:
\textbf{(i) cold start} (denoted as (0)), where the cache starts empty and is populated online by completed requests; and
\textbf{(ii) warm start} (denoted as (1k)), where the cache is initialized with 1K prompts and kept fixed during the test stream.

\textbf{Metrics.} We evaluate generation quality using \textbf{CLIP-SCORE}~\cite{hessel2021clipscore} and \textbf{VBench}~\cite{huang2024vbench}, where we report the average score (vbench-q) over four quality-related dimensions: subject consistency, background consistency, imaging quality, and motion smoothness.
For efficiency, we report \textbf{Latency(s)} (mean denoising time per request) and \textbf{Speedup} relative to the no-reuse baseline.
Chorus introduces a small preprocessing overhead on the order of hundreds of milliseconds.

\subsection{Overall Results on T2V Generation}
We evaluate Chorus on three Wan2.1 text-to-video variants (14B distilled 4-step, 14B 50-step, and 1.3B 50-step) in a serving-style inter-request reuse setting, with cold-start (0) and warm-start (1k) caches.
For each incoming request, Chorus retrieves a similar cached request and applies stage-wise reuse for generation.
Table~\ref{tab:main_experiments} summarizes generation quality (CLIP-SCORE, VBench-q) and efficiency (latency and speedup) compared with the no-reuse baseline, NIRVANA, and TeaCache.

\textbf{Distilled Model (4 steps).}
Experiments on Wan2.1-14B-T2V-distilled model demonstrate that Chorus significantly outperforms the inter-request baseline, NIRVANA, in terms of CLIP-Score and VBench-quality across both cold-start and warm-start scenarios, highlighting its superiority in generation quality. Regarding efficiency, despite the overhead introduced by the inclusion of an LLM and a segmentation model for quality enhancement, Chorus consistently achieves higher speedups and superior video quality compared to NIRVANA across all settings. Notably, it attains a speedup of 1.23× and ideally 1.45× speedup to the 4-step baseline, where intra-request caching is typically ineffective, validating its superiority in industrial serving scenarios.

\textbf{Vanilla Models (50 steps).}
Chorus demonstrates superior performance in the 50-step video diffusion model serving scenarios. Evaluated on both Wan2.1-14B and 1.3B, Chorus consistently achieves higher CLIP-Scores and VBench scores compared to NIRVANA, maintaining video quality competitive with the vanilla model. Notably, on the 14B model, Chorus even surpasses the vanilla model in VBench quality scores.
Furthermore, as our method is fully orthogonal to intra-request caching techniques, it can be seamlessly integrated with intra-request caching approaches like TeaCache. This combination yields speedups of up to 1.94$\times$ for the 14B model and 3.09$\times$ for the 1.3B model, with negligible degradation in visual quality.

\textbf{Key Insights.}
Three key observations emerge from our experiments.
(1)

Chorus exhibits a superior quality-efficiency trade-off compared to NIRVANA, most notably on the 4-step distilled model, where it simultaneously achieves higher acceleration and better generation quality metrics. Our approach effectively addresses the efficiency bottleneck in the few-step regime where intra-request caching is typically ineffective.
(2) On vanilla models, Chorus is orthogonal to intra-request caching: combining Chorus with TeaCache yields multiplicative speedups, highlighting the complementary nature of inter-request and intra-request reuse paradigms.
(3) Cache capacity is critical for performance, as larger caches naturally lead to higher hit rates and greater acceleration. With a sufficiently large cache, Chorus is projected to achieve peak speedups of 1.45$\times$ on the 4-step distilled model and up to 3.58$\times$ on vanilla models.

\subsection{Ablation Studies}
We ablate three key factors in Chorus: TGAA, SRD, and cache capacity.

All ablations are conducted on Wan2.1-14B-T2V-distilled.
To isolate component effects under cache-hit conditions, we construct a hit-only pool by filtering requests that have a valid match in the cache, and then randomly sample 150 requests from this pool for the TGAA/SRD ablations.
Overall, the ablations show that (1) TGAA effectively improve semantic alignment after reuse but is sensitive to the amplification strength; 
(2) SRD reduces denoising compute with a controllable speed--quality trade-off; and 
(3) larger caches improve reuse and yield larger gains.

\begin{table}[htbp]
\centering
\caption{\textbf{Ablation study of TGAA.} We fix Stage-1 reuse and toggle TGAA components (output vs. key amplification). Speedup is relative to the no-reuse baseline.}
\resizebox{\linewidth}{!}{
\begin{tabular}{lcccc}
\toprule
\textbf{Method} & \textbf{CLIP-SCORE$\uparrow$} & \textbf{vbench-q$\uparrow$} & \textbf{Latency(s)$\downarrow$} & \textbf{Speedup$\uparrow$} \\
\midrule
baseline                & 31.7581 & 0.8995 & 38.51 & 1.000x\\
Stage-1 only            & 30.5059 & 0.8988 & 28.20 & 1.365$\times$ \\
Stage-1 + TGAA (output) & 30.5883 & 0.9014 & 28.04 & 1.373$\times$ \\
Stage-1 + TGAA (key)    & 30.7958 & 0.8987 & 28.20 & 1.365$\times$ \\
Stage-1 + TGAA          & 30.7987 & 0.9013 & 28.20 & 1.365$\times$ \\

\bottomrule
\end{tabular}
}
\label{tab:TGAA-ablation}
\vspace{-2mm}
\end{table}

\begin{table}[htbp]
\caption{\textbf{Ablation study of SRD}. The $(m,n)$ controls mask expansion and the token compute fraction. Lower compute generally improves end-to-end speed with a small quality trade-off. 
Speedup is relative to the no-reuse baseline.}
\resizebox{\linewidth}{!}{
\begin{tabular}{lccccc}
\toprule
\textbf{Method} & \textbf{Compute(\%)$\downarrow$} & \textbf{CLIP-SCORE$\uparrow$} & \textbf{vbench-q$\uparrow$} & \textbf{Latency(s)$\downarrow$} & \textbf{Speedup$\uparrow$} \\
\midrule
Stage-1 + TGAA (w/o SRD) & 100.00 & 30.7987 & 0.9013 & 28.20 & 1.37$\times$ \\
Stage-1 + TGAA + SRD(12,20) & 90.91 & 30.7653 & 0.9013 & 27.13 & 1.42$\times$ \\
Stage-1 + TGAA + SRD(8,14) & 83.80 & 30.7469 & 0.9014 & 26.90 & 1.43$\times$ \\
Stage-1 + TGAA + SRD(0,0) & 38.42 & 30.3223 & 0.9015 & 24.13 & 1.60$\times$ \\
\bottomrule
\end{tabular}
}
\label{tab:ablation-SRD}
\vspace{-2mm}
\end{table}

\begin{figure}[ht]
    \centering
    \includegraphics[width=1\linewidth]{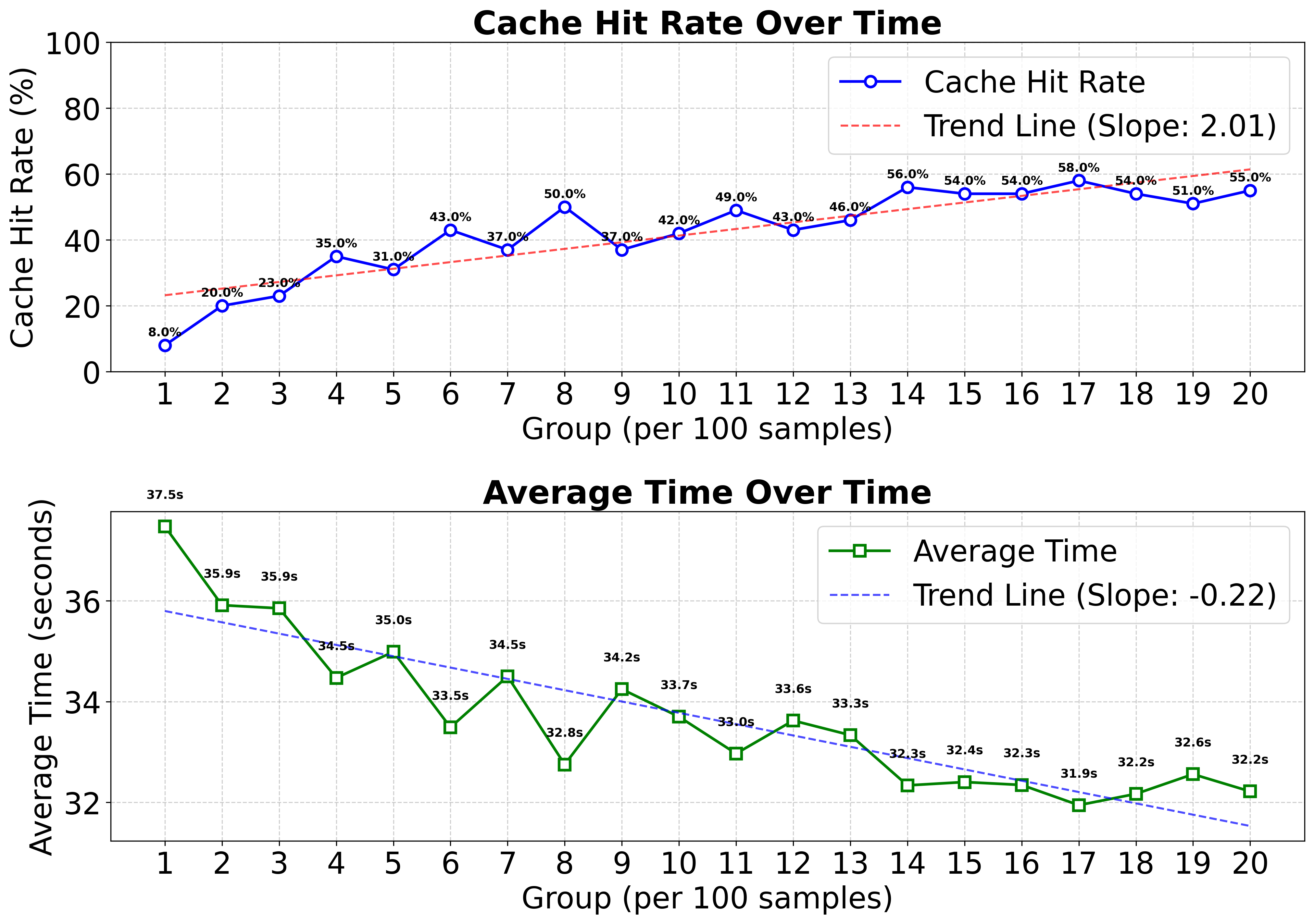}
    \vspace{-5mm}

    \caption{\textbf{Cache dynamics under cold start.}
Cache hit rate and average latency as the cache grows from an empty initialization.}
    \label{fig:cache_performance_trend}
    \vspace{-3mm}
\end{figure}

\begin{figure}
    \centering
    \includegraphics[width=1\linewidth]{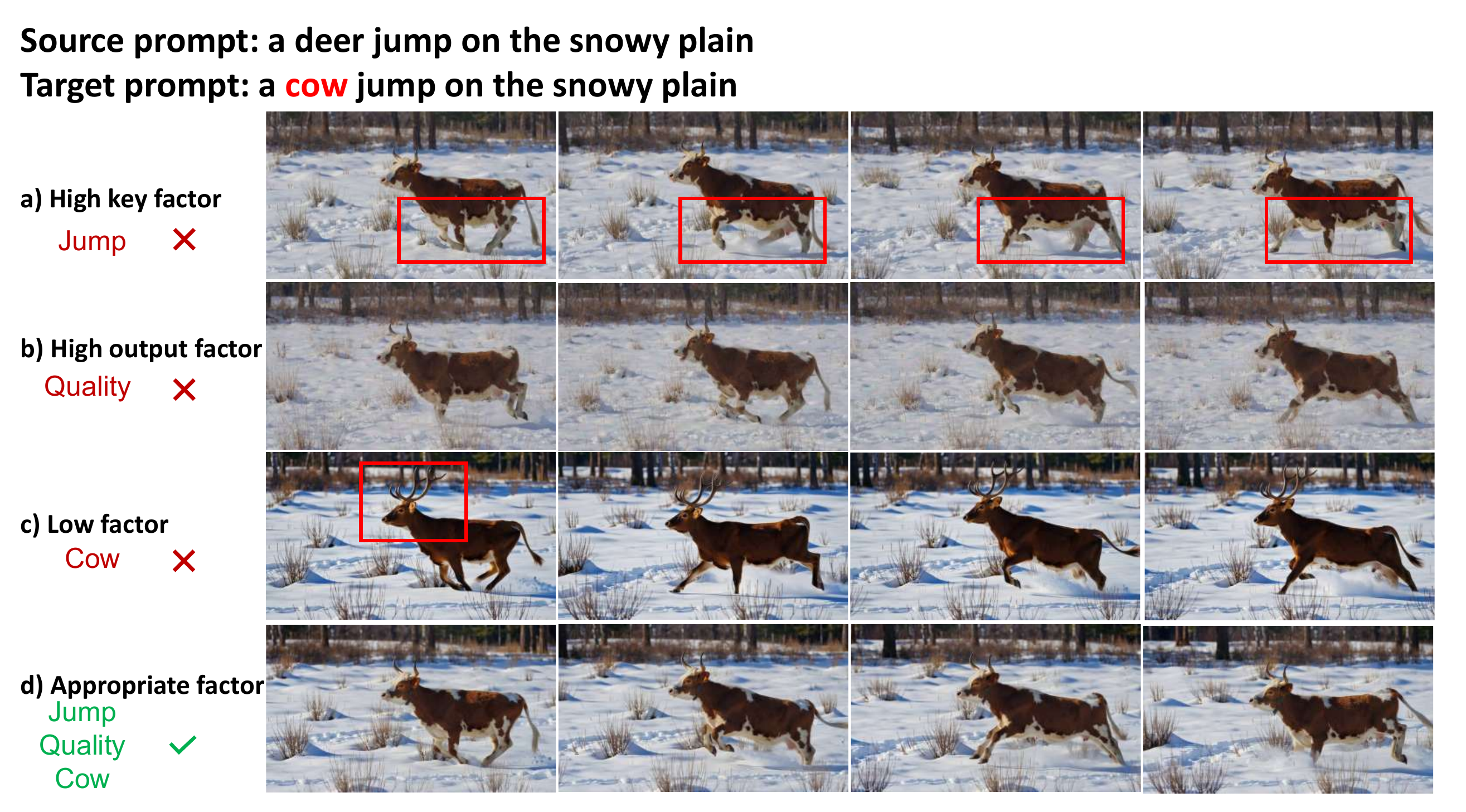}
    \vspace{-5mm}

    \caption{\textbf{Sensitivity of TGAA amplification factors.}
    (a) Over-amplified keys hurt motion. (b) Over-amplified outputs add artifacts. (c) Under-amplification fails to steer semantics.}
    \label{fig:cross-attention-enhancement ablation}
    \vspace{-5mm}
\end{figure}

\textbf{Ablation of TGAA.}
We keep Stage-1 reuse fixed and ablate TGAA by toggling output/key amplification under the same default strength schedule.
Table~\ref{tab:TGAA-ablation} quantifies the contribution of TGAA on top of Stage-1 reuse.
While Stage-1 alone yields a clear speedup, it also introduces semantic drift (CLIP-SCORE drops to 30.5059).
Adding TGAA mitigates this degradation with negligible overhead: output amplification improves CLIP-SCORE to 30.5883, and key amplification further boosts it to 30.7958.
Combining both achieves the best overall trade-off (30.7987 CLIP-SCORE and 0.9013 vbench-q), suggesting that the two mechanisms are complementary—output amplification stabilizes the generated content, whereas key amplification more directly strengthens target-condition binding. 

We further study sensitivity to the amplification factors by varying \(\gamma_k\) and \(\gamma_o\).
As shown in Fig.~\ref{fig:cross-attention-enhancement ablation}, TGAA is sensitive to the amplification strength and requires a dynamic decayed schedule across denoising steps. 
Over-amplifying keys factor (\(\gamma_k\)) over-focuses attention on conditional tokens, suppressing motion/attribute cues and causing motion artifacts or missing objects (Fig.~\ref{fig:cross-attention-enhancement ablation}a).
Over-amplifying outputs factor (\(\gamma_o\)) may inject artifacts and blur details (Fig.~\ref{fig:cross-attention-enhancement ablation}b).
Conversely, insufficient amplification fails to steer the generation toward the target concept (Fig.~\ref{fig:cross-attention-enhancement ablation}c).
These results show that TGAA is a lightweight, training-free inference-time enhancement, but its gains depend on reasonable choices of $(\gamma_k,\gamma_o)$ and a step-decay schedule to avoid over-steering.

\textbf{Ablation of SRD.} 
With Stage-1 reuse and TGAA fixed, we only vary SRD by changing the mask expansion parameters $(m,n)$, which directly controls the token compute fraction (Compute\%).
As shown in Table~\ref{tab:ablation-SRD}, lowering Compute\% consistently improves denoising speed, with a mild quality trade-off. 
The most aggressive setting SRD(0,0) reduces Compute\% to 38.4\% and achieves the largest speedup (1.60×), but also incurs the largest CLIP-SCORE drop. 
In contrast, moderate expansions such as SRD(8,14) and SRD(12,20) preserve quality (near-identical vbench-q and $\leq 0.05$ CLIP-SCORE drop) while still providing consistent acceleration ($1.42$--$1.43\times$).
We adopt SRD(8,14) as the default due to its favorable balance between efficiency and fidelity. A limitation is that SRD’s gains are bounded by mask coverage: when large regions must be denoised (e.g., heavy motion or broad appearance changes), Compute\% remains high and the speedup diminishes.

\textbf{Cache Growth and Hit-Rate Evolution.} 
During online serving from a cold start, the cache grows continuously, leading to a monotonic increase in hit rate and a steady reduction in latency (Fig.~\ref{fig:cache_performance_trend}). 
We report hit rate@0.75 and average latency aggregated over every 100 consecutive requests. 
As more requests are cached, the database better covers the prompt distribution, increasing the likelihood of finding a reusable match. 
Concretely, the hit rate remains below $20\%$ within the first 200 requests, but rises to $\sim 50\%$ after $\sim 1{,}500$ requests, corresponding to a $\sim 1.2\times$ speedup, with the trend still improving.
While our evaluation trace is not long enough to reach the high-hit-rate regime (e.g., $\geq 90\%$), these results should be viewed as conservative.
At scale, larger caches and workload locality can push the hit rate beyond $90\%$, driving speedup toward the $\sim 1.5\times$, where the benefit of inter-request reuse is most pronounced.

\vspace{-2mm}
\section{Conclusion}
\vspace{-1mm}
This paper presents Chorus, an inter-request caching reuse approach for accelerating video DiT serving at scale.
We introduce Token-Guided Attention Amplification mechanism to strengthen prompt conditioning and preserve semantic alignment, thereby enabling more aggressive full reuse and skipping additional denoising steps.
Next, we propose Selective Region Denoising to enable partial reuse during intermediate steps, further reducing computation.
Experiments show that Chorus substantially outperforms prior intra- and inter-request approaches, achieving higher speedups while preserving generation quality.
Notably, Chorus is applicable to industrial scenarios.
Even on industrial 4-step distilled models, where intra-request caching reuse becomes ineffective, Chorus delivers up to 45\% speedup.

\section*{Impact Statement}
This paper presents work whose goal is to advance the field of video DiT serving. 
There are many potential societal consequences of our work, none of which we feel must be specifically highlighted here.

\bibliography{example_paper}
\bibliographystyle{paperref}

\newpage
\appendix
\onecolumn
\section{More Visual Results}
\begin{figure} [H]
    \centering
    \includegraphics[width=1\linewidth]{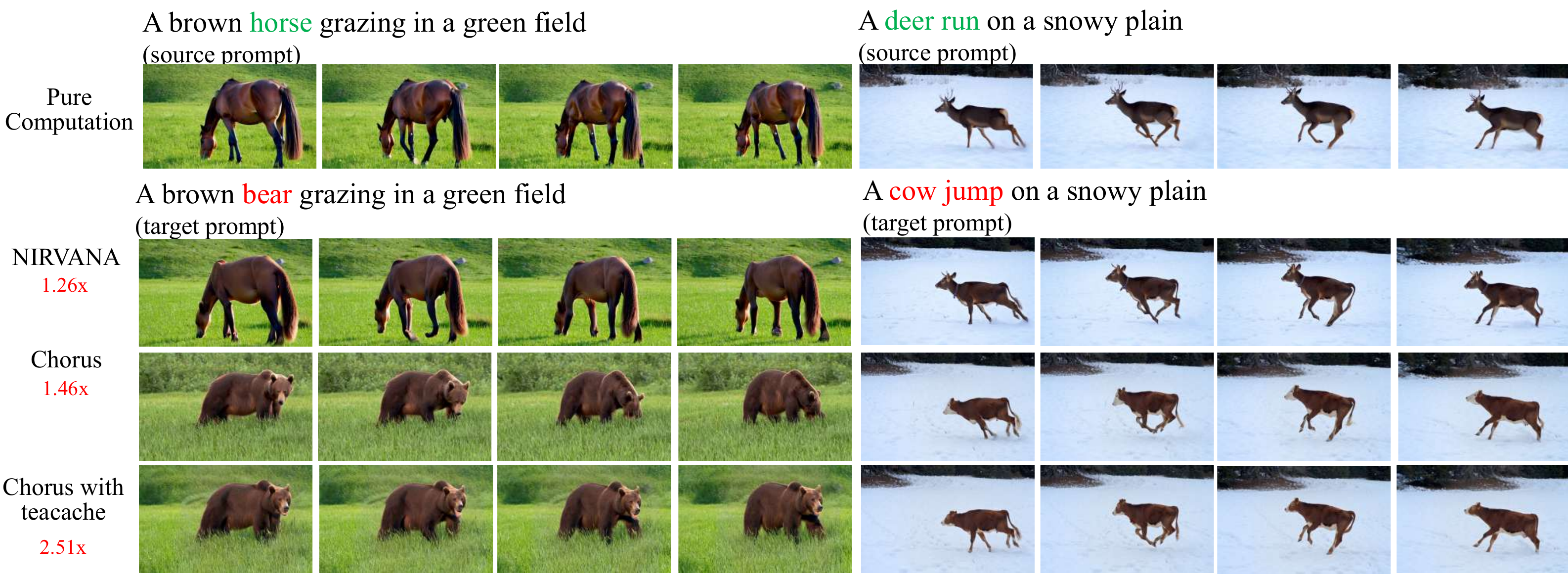}
    \caption{Results on Wan2.1-T2V-14B.}
    \label{fig:placeholder}
\end{figure}

\begin{figure} [H]
    \centering
    \includegraphics[width=1\linewidth]{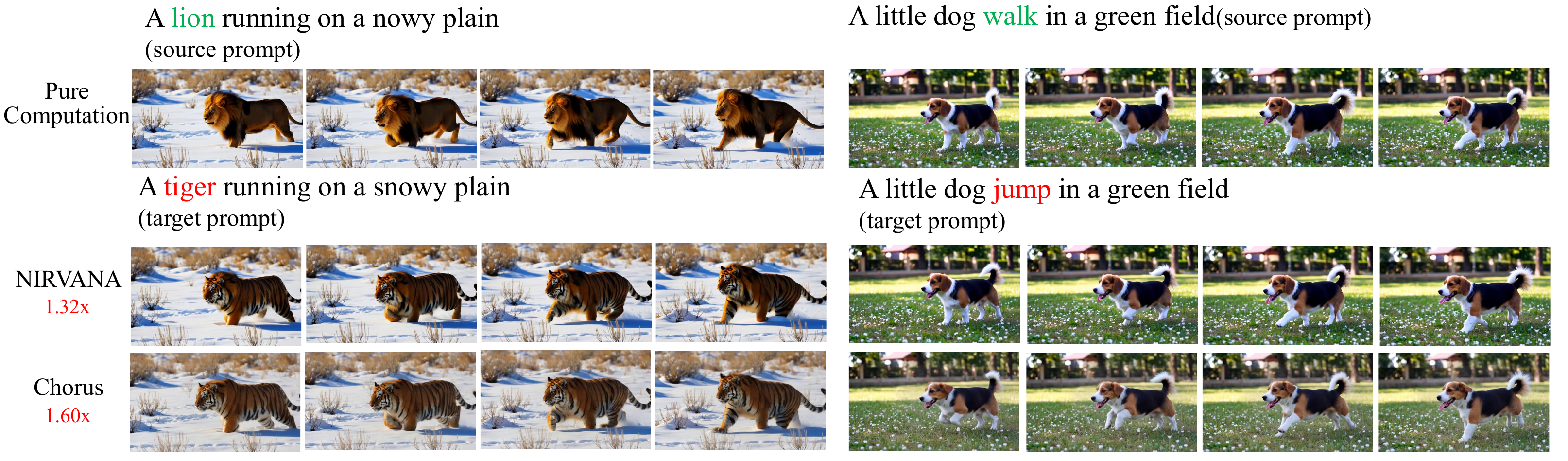}
    \caption{Results on distilled Wan2.1-T2V-14B}
    \label{fig:visual-result-2}
\end{figure}

\section{Experimental Details}
\label{sec:appendix_experiments}

\subsection{Dataset Processing}
We evaluate Chorus on VidProM~\cite{wang2024vidprom-dataset}, which contains 1.67M unique text-to-video prompts. 
To improve data quality, we filter out low-quality prompts using a simple length-based heuristic, resulting in a subset of 50{,}000 prompts.
To accelerate evaluation while preserving workload locality, we embed all remaining prompts using the CLIP text encoder (768-d) and apply $k$-means clustering~\cite{lloyd1982least} with $k=20$ on the normalized embeddings. 
We then select one cluster with approximately 2{,}000 prompts as our evaluation workload. 
From this cluster, we randomly sample 1{,}000 prompts to initialize the cache (index), and use the remaining 1{,}000 prompts as the test stream for video generation and metric evaluation.

\subsection{Cache Backend and Retrieval}
We use ChromaDB as the cache backend, which provides lightweight vector similarity search. 
Each prompt is represented by a 768-d CLIP text embedding, and we retrieve the top-1 nearest neighbor under cosine similarity. 
With a cache size of 1{,}000 entries, retrieval typically takes on the order of tens of milliseconds, which is negligible compared to the end-to-end denoising latency.

\end{document}